\documentclass{article}
\usepackage{hyperref}
\usepackage{spconf}
\usepackage{graphicx}
\usepackage{caption}
\usepackage{subcaption}
\usepackage{algorithm,algorithmicx}
\usepackage[noend]{algpseudocode}
\usepackage{amsmath}
\usepackage[capitalise]{cleveref}
\usepackage{todonotes}
\usepackage{enumitem}
\usepackage{eqnarray}

\title{Streaming Simultaneous Speech Translation \\ with Augmented Memory Transformer}

\name{
    Xutai Ma$^{\dagger}$ \qquad
    Yongqiang Wang$^{\star}$ \qquad
    Mohammad Javad Dousti$^{\star}$ \qquad
    Philipp Koehn$^{\dagger\star}$ \qquad
    Juan Pino$^{\star}$
}
\address{
      $^{\star}$ Facebook \\
      $^{\dagger}$ Johns Hopkins University
}

\begin{document}

\maketitle

\begin{abstract}
Transformer-based models have achieved state-of-the-art performance on speech translation tasks.
However, the model architecture is not efficient enough for streaming scenarios
since self-attention is computed over an entire input sequence
and the computational cost grows quadratically
with the length of the input sequence.
Nevertheless, most of the previous work on simultaneous speech translation,
the task of generating translations from partial audio input,
ignores the time spent in generating the translation when analyzing the latency.
With this assumption, a system may have good latency quality trade-offs but be inapplicable in real-time scenarios.
In this paper,
we focus on the task of \emph{streaming simultaneous speech translation},
where the systems are not only capable of translating with partial input but are also able to handle very long or continuous input.
We propose an end-to-end transformer-based sequence-to-sequence model,
equipped with an \textit{augmented memory transformer encoder},
which has shown great success on the streaming automatic speech recognition task
with hybrid or transducer-based models.
We conduct an empirical evaluation of the proposed model on segment, context and memory sizes and we compare our approach to a transformer with a unidirectional mask. 
\footnote{This work has been submitted to the IEEE for possible publication. Copyright may be transferred without notice, after which this version may no longer be accessible.}
\end{abstract}

\section{Introduction}
Streaming speech translation targets low latency scenarios such as simultaneous interpretation.
Unlike the streaming automatic speech recognition (ASR) task
where input and output are monotonically aligned, translation needs a larger future context due to reordering.
Generally, simultaneous translation models start to translate with partial input,
and then alternate between generating predictions and consuming additional input.
While most previous work on simultaneous translation focus on text input,
\cite{cho2016can,gu2017learning,ma-etal-2019-stacl, arivazhagan-etal-2019-monotonic, ma2019monotonic}
the end-to-end approach for simultaneous speech translation
has also very recently attracted interest from the community
\cite{ren2020simulspeech, ma2020simulmt}
due to potentially lower latency compared with cascade models~\cite{oda2014optimizing,fugen2007simultaneous, fujita2013simple, muller2016lecture}.
However, most studies tend to focus on an ideal setup,
where the computation time to generate the translation is neglected.
This assumption may be reasonable for text to text but not for speech to text translation since the latter has much longer input sequences.
A simultaneous speech translation model may have the ability to generate translations with partial input but may not be useful for real-time applications because of slow computation in generating output tokens.

While the majority of previous work on streaming speech to text tasks has focused on ASR, most prior work on ASR is not directly applicable to translation.
Encoder-only or transducer structures are widely implemented,
since ASR assumes the output is monotonically aligned to the input.
In order to achieve an efficient streaming simultaneous speech translation model, we combine streaming ASR and simultaneous translation techniques and introduce an end-to-end transformer-based speech translation model with an augmented memory encoder~\cite{wu2020streaming}.

The augmented memory encoder has shown considerable improvements
on latency with little sacrifice on quality with hybrid or transducer-based models on the ASR task.
It incrementally encodes fixed-length sub-sentence level segments
and stores the history information with a memory bank,
which summarizes each segment.
The self-attention is only performed on the current segment and memory banks.
A decoder with simultaneous policies is then introduced on top of the encoder.
We apply the proposed model to the simultaneous speech translation task
on the MuST-C dataset \cite{di-gangi-etal-2019-must}.

This paper is organized as follows.
We first define the evaluation method for simultaneous speech translation. We then introduce the model based on the augmented memory transformer. Finally, we conduct a series of experiments to demonstrate the effectiveness of the proposed approach.

\section{Evaluation}
This paper focuses on streaming \emph{and} simultaneous speech translation,
which features two additional capabilities compared to a traditional offline model.
The first one is the efficient computation needed to handle streaming input, and the second is the ability to start a translation with partial input, then
dynamically generate additional tokens or read additional input.
Both factors need to be considered for evaluation.

A simultaneous system is evaluated with respect to quality,
usually with BLEU, and latency.
Latency is evaluated with computation-aware and non computation-aware Average Lagging (AL)~\cite{ma2020simulmt, ma-etal-2019-stacl}.
Denote the input sequence as $\mathbf{X}=[\mathbf{x}_1,...]$,
where each element is a feature vector extracted from a sliding window of size $T$
and the translation of the system $\mathbf{Y}=[y_1,...]$ and a reference translation $\mathbf{Y}^*$.
Non computation-aware (NCA) latency is defined as follows
\begin{equation}
    \text{AL} = \frac{1}{\tau(|\mathbf{X}|)} \sum^{\tau(\mathbf{X})}_{i=1} d(y_i) - \frac{|\mathbf{X}|}{|\mathbf{Y}^*|}  \cdot T \cdot (i - 1)
    \label{eq:al}
\end{equation}
where $\tau(|X|)$ is the index of the first target token
when the system has read the entire source input,
and $d(y_i)$ is the duration of the speech that has been read when generating word $y_i$.
Additionally, a computation-aware (CA) version of AL is also considered,
by replace $d(y_i)$ with the time needed to generate $y_i$.

\section{Model}
\begin{figure}[]
    \centering
    \includegraphics[width=0.4\textwidth]{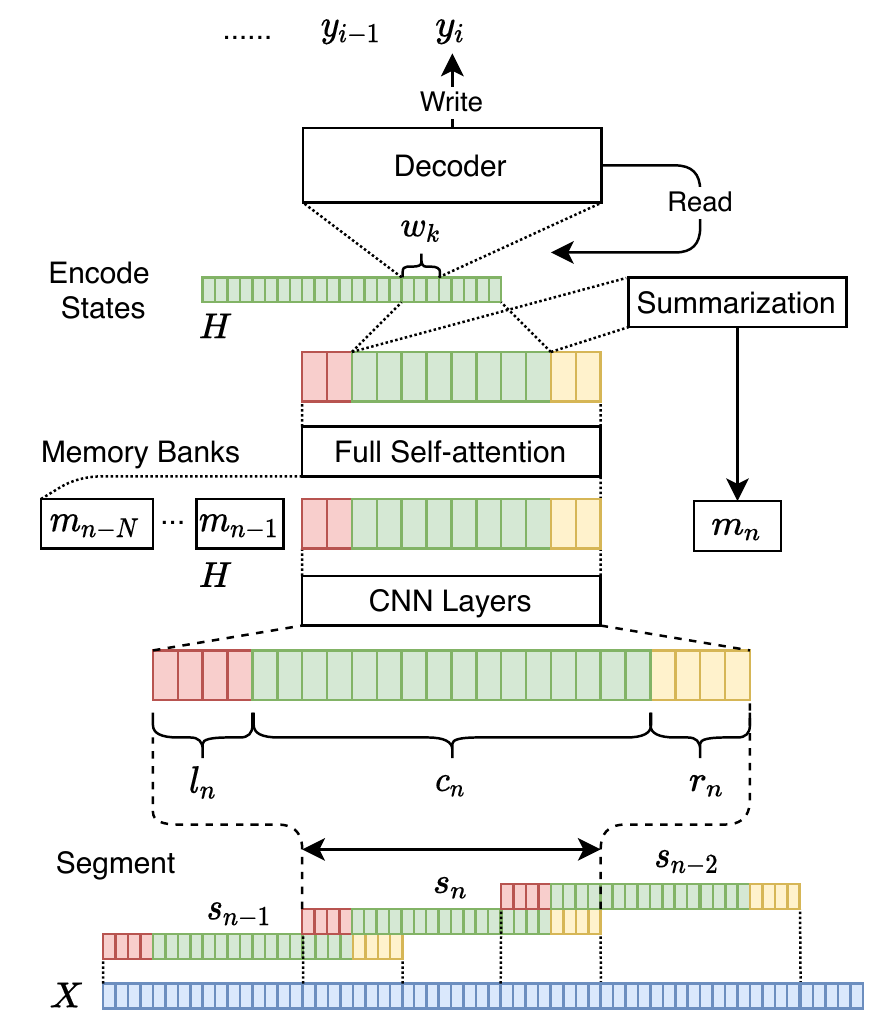}
    \caption{Architecture of streaming transformer model with an augmented memory encoder.}
    \label{fig:model}
\end{figure}

The proposed streaming speech translation model, illustrated in \cref{fig:model},
consists of two components,
an augmented memory encoder and a simultaneous decoder.
The encoder incrementally and efficiently encodes streaming input,
while the decoder starts translation with partial input,
then interleaves reading new input and predicting target token
under the guidance of a simultaneous translation policy.


\subsection{Augmented Memory Encoder}
The self-attention module in the original transformer model \cite{vaswani2017attention} attends to the entire input sequence,
which precludes streaming capability.
Denote $\mathbf{H} = [\mathbf{h}_1, ..]$ the input of a certain encoder layer.
Each self-attention projects the input into query, key and value.
\begin{equation}
    \mathbf{Q} = W_q \mathbf{H}, \mathbf{K} = W_k \mathbf{H}, \mathbf{V} = W_v \mathbf{H}
\end{equation}
At each position $j$, a weight is calculated as follows
\begin{equation}
    \alpha_{jj'} = \frac{\text{exp}(\beta \cdot \mathbf{Q}^T_j \mathbf{K}_{j'})}{\sum_{k} \text{exp}(\beta \cdot \mathbf{Q}^T_j \mathbf{K}_{k})}
\end{equation}
The self-attention at position $j$ can then be calculated as
\begin{equation}
    \mathbf{Z}_j = \sum_{j'} \alpha_{jj'} \mathbf{V}_{j'}
\end{equation}
The calculation of self-attention make it inefficient for streaming applications.
\cite{wu2020streaming} proposes an augmented memory transformer encoder to address this issue.
Instead of attending to entire input sequence $\mathbf{X}$,
the self-attentions are applied on a sequence of sub-utterance level segments $\mathbf{S}=[\mathbf{s}_1,...]$.
A segment $\mathbf{s}_n$, which contains a span of input features, consists of three parts:
left context $\mathbf{l}_n$ of size $L$,
main context $\mathbf{c}_n$ of size $C$ and right context $\mathbf{r}_n$ of size $R$.
Each segment overlaps with adjacent segments ---
the overlap between current and previous segment is $\mathbf{l}_n$,
and between current and the next segment is $\mathbf{r}_n$.
Self-attention is computed at the segment level, which reduces the amount of computation.
The new query, key and value for each segment are
\begin{align}
    \mathbf{q}_n &= \mathbf{W_q} (\mathbf{l}_n, \mathbf{c}_n, \mathbf{r}_n, \sigma_n) \\
    \mathbf{k}_n &= \mathbf{W_k} (\mathbf{M}_{n-N:n-1}, \mathbf{l}_n, \mathbf{c}_n, \mathbf{r}_n) \\
    \mathbf{v}_n &= \mathbf{W_v} (\mathbf{M}_{n-N:n-1}, \mathbf{l}_n, \mathbf{c}_n, \mathbf{r}_n)
\end{align}
Where $\sigma_n = \sum_{\mathbf{x}_k \in \mathbf{s}_n} \mathbf{x}_k$ is a summarization of the segment $\mathbf{s}_n$,
and $\mathbf{M}_{n-N:n} = [\mathbf{m}_{n-N}, ..., \mathbf{m}_{n - 1}]$ are the memory banks.
Each memory bank is calculated as follows:
\begin{equation}
    \mathbf{m}_n = \sum_{j'} \alpha_{-1, j'} (\mathbf{v}_n)_{j'}
\end{equation}
which is introduced to represent history information.
A hyperparameter $N$ controls how many memory banks are retained.
The self-attention is then calculated as
follows:
\begin{align}
    \alpha_{jj'} &= \frac{\text{exp}(\beta \cdot (\mathbf{q}^T_n)_j (\mathbf{k}_n)_{j'})}{\sum_{k} \text{exp}(\beta \cdot (\mathbf{q}^T_n)_j (\mathbf{k}_n)_{k})} \\
    (\mathbf{z}_n)_j &= \sum_{j'} \alpha_{j, j'} (\mathbf{v}_n)_{j'} \\
    \text{where } &N + L < j \le N + L + C
\end{align}
Then only the central encoder states are kept and a the concatenation of the segment states
$\mathbf{Z} = [\mathbf{z}_1, ...]$ is passed to decoder.
Because of the left and right contexts,
an arbitrary encoder can run on the segments without boundary mismatch.
In this paper, we adapt the encoder of the convtransformer architecture~\cite{inaguma-etal-2020-espnet}.
The encoder first consists of convolutional layers with stride 2 that subsample the input.
Full self-attention layers can then be calculated.

\subsection{Simultaneous Decoder}
A simultaneous decoder starts translation with partial input
based on a policy.
Simultaneous policies decide whether the model should read new inputs
or generate a new prediction at a given time.
However, different from text translation,
our preliminary experiments show that for simultaneous speech translation,
encoder states are too granular for policy learning.
Thus, we adopt the idea of pre-decision \cite{ma2020simulmt}
for better efficiency by making simultaneous read and write decision
on chunks of encoder states.
Here, the simpler fixed pre-decision strategy is used where the decision is made every fixed number of encoder states.
Denote the sequence of chunks are $\mathbf{W}=[\mathbf{w}_1,...]$
and the start and end encoder state index of $\mathbf{w}_k$ is $W_s(k), W_e(k)$.
Denote the prediction of model $\mathbf{Y}=[y_1,...]$,
the general decoding algorithm of
a simultaneous policy $\mathcal{P}$
with augmented memory transformer is described in \cref{alg:simul decoding}.
\begin{algorithm}[h]
\small
  \caption{Chunk-based simultaneous policy with an augmented memory encoder}
  \label{alg:simul decoding}
  \begin{algorithmic}[1]
  \Require{Chunk-based simultaneous policy $\mathcal{P}$}
  \Require{Streaming input $\mathbf{X}$. Memory banks $\mathbf{M}$. Prediction $\mathbf{Y}$}
  \Require{Maximum memory size $N$. Decision chunk size $W$}
  \Require{Central context size $C$. Encoder pooling ratio $R$}
  \Require{$i=1, n=1, k=1$.}
  \Require{$W_e(1)=1, y_0=\text{BOS}$ }
  \While{$y_{i-1} \neq \text{EndOfTranslation}$}
        \If{$W_e(k) + W > n \cdot C \cdot R$}
            \State $\mathbf{z}_n, \mathbf{m}_n = \text{Encoder}(\mathbf{s}_n, \mathbf{M}_{n-N:n-1})$
            \State $\mathbf{Z} = [\mathbf{Z}, \mathbf{z_n}], \mathbf{M} = [\mathbf{M}, \mathbf{m_n}]$
            \State $n = n + 1$ \, $\#$ Read a new segment of input features
        \EndIf
        \State $\mathbf{w}_k = \text{Summarize}(\mathbf{Z}_{W_s(k):W_e(k)})$
        \State $p_{ik} = \mathcal{P}([\mathbf{Y}_{1:i-1}], \mathbf{w}_k)$
            \If{$p_{ik} > 0.5$}
                \State $y_{i} = \text{Decoder}([\mathbf{Y}_{1:i-1}], \mathbf{Z})$
                \State $i=i+1$ \, $\#$ Predict a target token
            \Else
                \State $W_s(k + 1) = W_e(k) + 1$
                \State $k = k + 1$ \,$\#$ Move to the next chunk of encoder states
            \EndIf
    \EndWhile
  \end{algorithmic}
\end{algorithm}
In theory,
\Cref{alg:simul decoding} supports arbitrary simultaneous translation policies.
In this paper, for simplicity, wait-$k$~\cite{ma-etal-2019-stacl} is used.
It waits for $k$ source tokens
and then operating then reading and writing alternatively.
Notice that our method is compatible with an arbitrary simultaneous translation policy.

Note that the decoder self-attention still has access to
all previous decoder hidden states; in order to preserve streaming capability for the decoder,
decoder states are reset every time an end-of-sentence token is predicted.
The augmented memory is not introduced in the decoder because the
target sequence is dramatically smaller than the source speech sequence.
The decoder can still predict a token in a negligible time
compared with encoding source with the input becoming longer.
\section{Experiments}
\label{sec:experiments}
Experiments were conducted on the English-German MuST-C dataset \cite{di-gangi-etal-2019-must}.
The training data consists of 408 hours of speech and 234k sentences of text.
We use Kaldi~\cite{povey2011kaldi} to extract 80 dimensional log-mel filter bank features.
The features are computed with a 25$ms$ window size and a 10$ms$ window shift and normalized with global cepstral mean and variance.
Text is tokenized with a SentencePiece\footnote{\url{https://github.com/google/sentencepiece}} 10k unigram vocabulary. 
Translation quality is evaluated with case-sensitive detokenized BLEU with \textsc{SacreBLEU}\footnote{\url{https://github.com/mjpost/sacrebleu}}.
The latency is evaluated by Average Lagging~\cite{ma-etal-2019-stacl, ma2020simulmt}, with the SimulEval toolkit\footnote{\url{https://github.com/facebookresearch/SimulEval}}.

The speech translation model is based on the convtransformer architecture~\cite{inaguma-etal-2020-espnet}.
It first contains two convolutional layers with subsampling ratio of 4.
Both encoder and decoder have a hidden size of 256 and 4 attention heads.
There are 12 encoder layers and 6 decoder layers.
The model is trained with label smoothed (0.1) cross entropy.
We use the Adam optimizer~\cite{adam}, with a learning rate of 0.0001 and an inverse square root schedule.

We use a simplified version of \cite{ren2020simulspeech} as our baseline model.
A unidirectional mask is introduced to prevent the encoder from looking into future information.
For baseline models,
we follow common practice for simultaneous text translation where the entire encoder is updated once there is new input.
Instead of a multi-task setting and a decision making process depending on word boundaries obtained from the auxiliary ASR task, we make decisions on a fixed size chunk of encoder states, following \cite{ma2020simulmt}. Our choice is motivated by the fact that in \cite{ma2020simulmt}, a fixed chunk size gave similar quality-latency trade-offs as word boundaries.

All transformer-based speech translation models are first pre-trained on the ASR task, in order to initialize the encoder.
Each experiment is run on 8 Tesla V100 GPUs with 32 GB memory.
The code is developed based on Fairseq\footnote{\url{https://github.com/pytorch/fairseq}},
and it will be published upon acceptance.
All scores are reported on the dev set.

\section{Results}
We first analyze the effect of the segment and context sizes, and use the resulting optimal settings for further analysis on the maximum number of memory banks and for comparison with the baseline.
\subsection{Effect of Segment and Context Size}
\begin{figure}[h]
    \centering
    \includegraphics[width=0.45\textwidth]{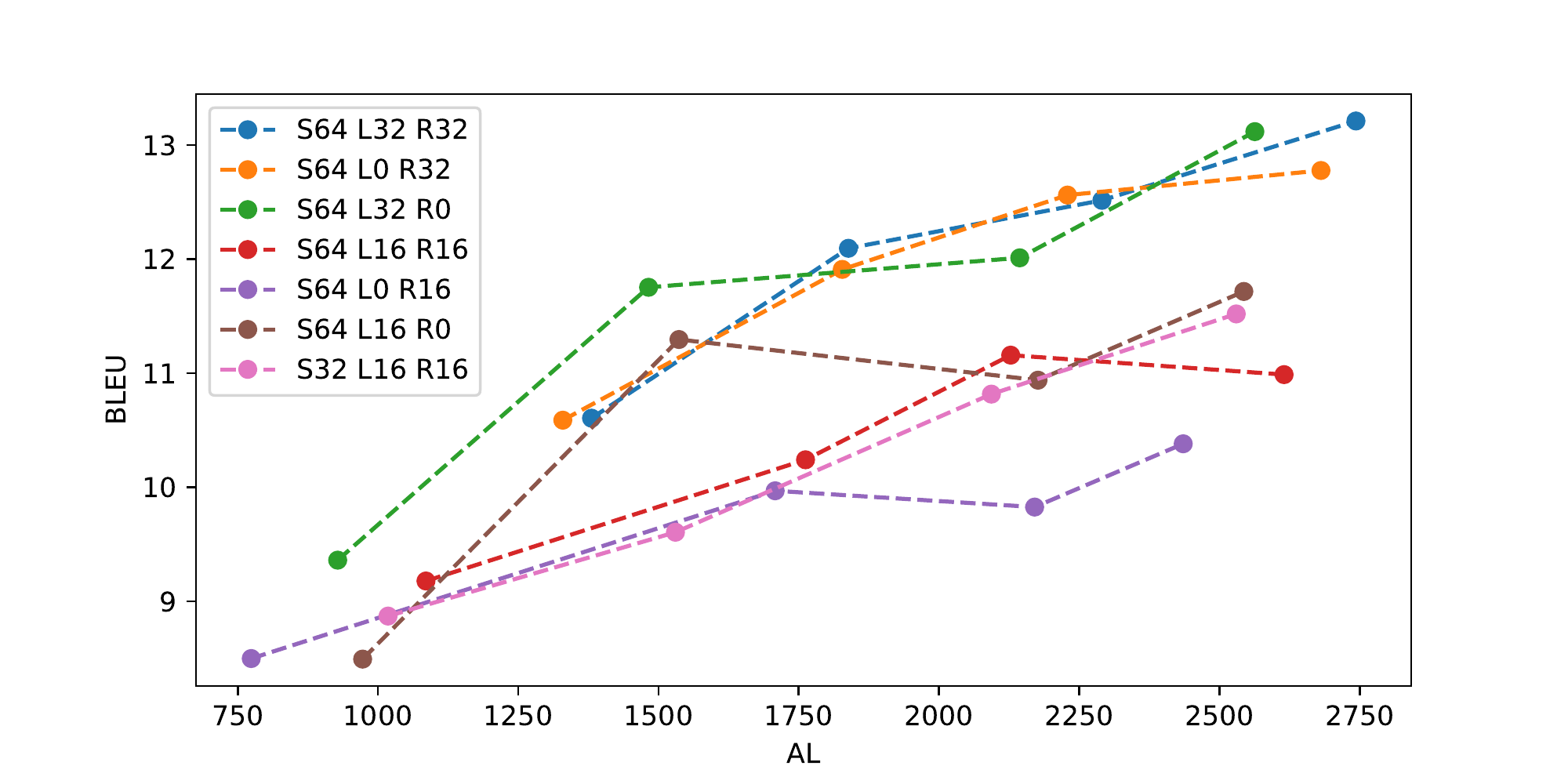}
    \caption{
        Effect of segment, left and right context size.
        Each curve represents wait-$k$, $k=1, 3,5,7$ policies.
        The size is measured on a frame of 10 ms.
        ``S$\{x\}$ L$\{y\}$ R$\{z\}$'' means a encoder with segment size $x$,
        left size $y$ and right size $x$.
    }
    \label{fig:segment_size}
\end{figure}

We analyze the effect of different segment, left and right context sizes.
For all experiments,
we use the wait-$k$ ($k=1,3,5,7$) policy on a chunk of 8 encoder states~\cite{ma2020simulmt}.
The latency-quality trade-offs with different sizes are shown in \Cref{fig:segment_size}.
We first observe that,
increasing the left and right context size will improve the quality with very small trade-off on latency, 
for instance, from curve ``S64 L16 R16'' to ``S64 L32 R32''.
This indicates that context of both sides can alleviate the boundary effect.
We also notice that when we reduce the segment size from 64 to 32,
the BLEU score decreases dramatically.
Similar observations are made in \cite{wu2020streaming} but the ASR models are more robust to decreasing the segment and context sizes. We hypothesize that reordering in translation makes the model more sensitive to these sizes.
\setlength{\belowcaptionskip}{-10pt}
\subsection{Number of Memory Banks}
\begin{figure}[h!]
    \centering
    \includegraphics[width=0.45\textwidth]{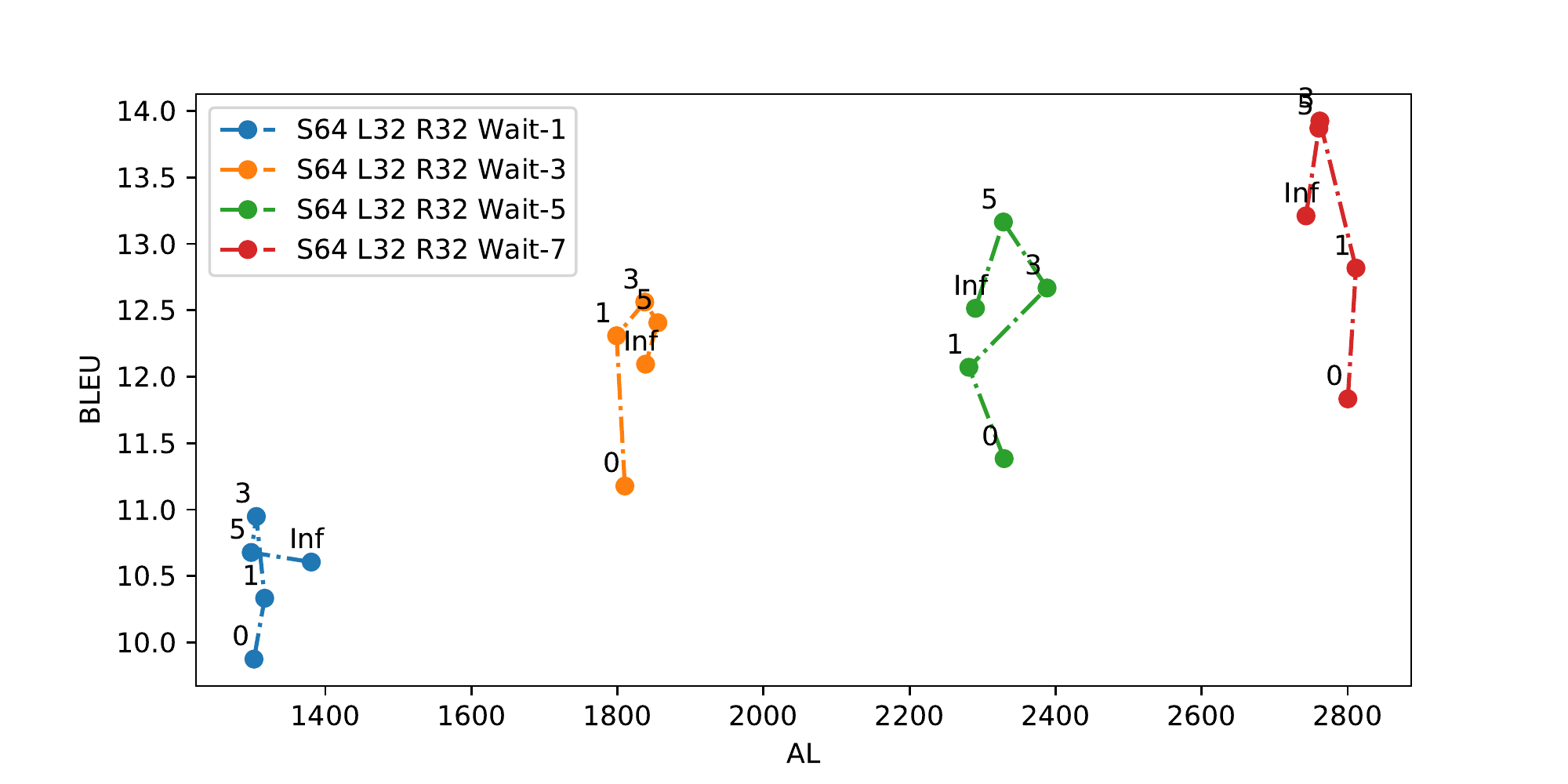}
    \caption{
        Effect of the maximum number of memory banks.
        Each curve represents one policy.
        The number on top of the nodes indicates the number of memory banks being kept.
    }
    \label{fig:memory_size}
\end{figure}

In streaming translation, the input is theoretically infinite.
In order to prevent memory explosion, we explore the effect of reducing the number of the memory banks.
\cref{fig:memory_size} shows the effect of different numbers of memory banks.
We can see that the model is very robust to the size of the memory banks.
Similar to \cite{wu2020streaming},
when the maximum number of memory banks is large, for instance, larger than 3, there is little or no performance drop.
However, we still observe a drop in performance with a maximum number of one memory bank. Finally, we found that training with different maximum numbers of memory banks was necessary as limiting the number of memory banks only at inference time degraded performance.

\subsection{Comparison with Baseline}
\begin{figure}[h]
    \centering
    \includegraphics[width=0.45\textwidth]{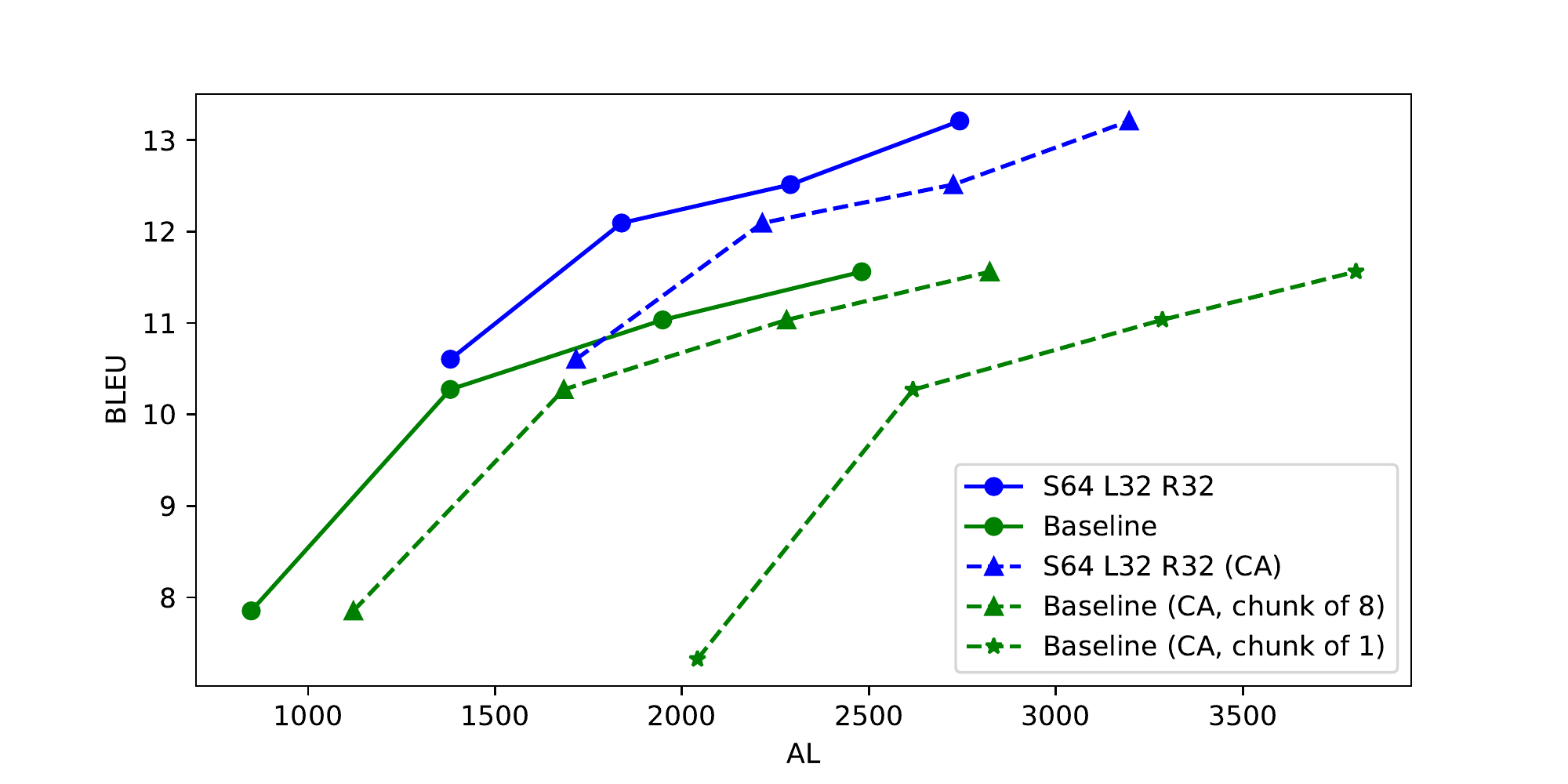}
    \caption{
        Comparison with baseline model.
        CA indicates measured by computation aware latency. Chunk of $x$ means that the encoder states are updated every $x$ steps.
    }
    \label{fig:baseline}
\end{figure}

In \cref{fig:baseline}, we compared our model with the baseline model described in \cref{sec:experiments}.
The proposed model achieves better quality with an increase in computation aware and non computation aware latency.
The baseline achieves competitive latency because it only updates encoder states every 8 steps. However, there may be instances where recomputing encoder states every step may be needed, for example in the case of a flexible pre-decision module
or when a the model includes a boundary detector~\cite{ren2020simulspeech}.
We can see in \cref{fig:baseline} that the computation aware AL for the baseline increases substantially with a chunk of size 1.

\section{Conclusion}
In this paper, we tackle the real-life application of streaming simultaneous speech translation. We propose a transformer-based model, equipped with an augmented memory, in order to handle long or streaming input. We study the effect of segment and context sizes, and the maximum number of memory banks. We show that our model has better quality with an acceptable latency increase compared with a transformer with unidirectional mask baseline and presents better quality-latency trade-offs than that baseline where encoder states are recomputed at every step.

\bibliographystyle{IEEEbib}
\bibliography{bibliography/simultaneous_translation,bibliography/streaming_asr,bibliography/machine_translation}

\end{document}